\documentclass[runningheads]{llncs}
\usepackage[T1]{fontenc}
\usepackage{graphicx}
\usepackage{booktabs}
\usepackage{amsmath}
\usepackage{amssymb}
\usepackage{xurl}
\usepackage[misc]{ifsym}
\newcommand{\corr}{(\Letter)}
\usepackage{hyperref}

\begin{document}
\title{From Clinical Risk Prediction to Operational Decisions: An Uncertainty-Aware Framework for ICU Staffing}

\titlerunning{An Uncertainty-Aware Framework for ICU Staffing}
\author{Jiayan Yang\inst{1}\corr \and
Ad Feelders\inst{1} \and
Arno Siebes\inst{1} 
}
\authorrunning{J. Yang et al.}
\toctitle{From Clinical Risk Prediction to Operational Decisions: An Uncertainty-Aware Framework for ICU Staffing}
\tocauthor{Jiayan Yang, Ad Feelders, Arno Siebes}

\institute{
Utrecht University, Princetonplein 5, 3584 CC Utrecht, the Netherlands \\ \email{\{j.yang2, A.J.Feelders, A.P.J.M.Siebes\}@uu.nl}
}

\maketitle              

\begin{abstract}
Advances in electronic health records (EHR) and large-scale critical care databases have enabled the development of machine learning models for the early detection of patient deterioration. While these models often achieve strong predictive performance for outcomes such as sepsis and mortality, most studies evaluate them purely on discrimination metrics. This leaves a critical gap: how to systematically translate probabilistic clinical risk estimates into downstream operational decisions under real-world constraints, such as shift rigidity and hiring delays.

This study proposes an uncertainty-aware anticipatory staffing framework that bridges short-term patient risk prediction and staff planning without altering the upstream predictive model. Instead of relying on naive deterministic aggregation or fixed safety buffers, the framework explicitly models dynamic predictive uncertainty and evaluates extreme shortage exposure using Conditional Value-at-Risk (CVaR). These aggregated, risk-calibrated demand signals are then embedded into an hourly rolling-horizon scheduling policy to generate delay-constrained staffing decisions.

Evaluated on realistic trajectories from the MIMIC-IV database, the proposed framework demonstrates a favorable cost-shortage trade-off compared to deterministic and purely reactive baselines. Pareto frontier analysis illustrates how uncertainty-aware demand construction changes resource allocation under varying safety preferences, while stress testing confirms the framework's operational robustness even when subjected to systematic predictive bias in the form of calibration drift. By formalizing this translation layer, this work provides a principled, transparent pathway for safely integrating probabilistic AI outputs into high-stakes healthcare operational systems.

\end{abstract}

\section{Introduction}

The increasing adoption of electronic health records (EHRs) has created new opportunities for data-driven analysis and decision support in healthcare. Machine learning models trained on large-scale clinical datasets are now widely used to predict adverse outcomes such as sepsis onset \cite{nemati2018interpretable,moor2020predicting}, patient deterioration or unplanned intensive care unit (ICU) transfer \cite{churpek2016multicenter}, and in-hospital mortality \cite{rajkomar2018scalable}. Publicly available databases such as MIMIC-IV have further accelerated research by providing large de-identified patient cohorts for model development and evaluation \cite{johnson2023mimic}. These models increasingly enable early identification of high-risk patients.

Despite these advances, predictive outputs are rarely translated into operational decisions. In practice, hospital resource planning often relies on historical workload averages or deterministic demand estimates \cite{kim2015two,green2007coping}, while the probabilistic information produced by predictive models remains underutilized.

A key challenge lies in the mismatch between prediction and decision levels. Clinical prediction models typically estimate risk at the patient level, whereas operational decisions such as staffing planning should be made at the population level. Translating individual risk predictions into reliable estimates of aggregate demand therefore remains a critical but underexplored problem.

Recent work has proposed integrating prediction and decision-making through predict-then-optimize frameworks \cite{elmachtoub2022smart} and decision-focused learning approaches \cite{shah2022decision}. These methods jointly optimize predictive and decision components within an end-to-end framework \cite{kotary2021end}. However, they typically require retraining predictive models and differentiable optimization layers. In clinical practice, predictive models are often deployed as standalone systems under regulatory and operational constraints, making such modifications difficult once models are validated and in use \cite{wu2021how,sendak2020presenting}.

In this work, we study how patient-level deterioration predictions can be translated into actionable signals for ICU staffing planning without modifying the upstream predictive model. We propose an uncertainty-aware framework that aggregates probabilistic deterioration risks across patients to estimate population-level demand, which is then incorporated into an optimization-based staffing model that explicitly balances staffing cost and operational risk.

We evaluate the framework using real-world ICU data from MIMIC-IV under multiple demand scenarios. Results show that probabilistic aggregation produces more stable and interpretable staffing demand estimates compared with deterministic baselines.

The main contributions of this work are: (i) identifying the challenge of translating patient-level deterioration predictions into population-level demand signals for ICU staffing planning; (ii) proposing an uncertainty-aware aggregation framework that converts predictive risks into stochastic demand estimates; and (iii) demonstrating improved stability and interpretability of staffing decisions through empirical evaluation on real-world clinical data.
\section{Related Work}
We review three related research areas: clinical risk prediction using machine learning, optimization methods for healthcare resource planning, and recent approaches that integrate predictive models with operational decision-making.
\subsection{Machine Learning for Clinical Risk Prediction}

Early identification of clinical deterioration has been widely studied in clinical research and medical informatics, as timely detection enables early intervention and prevents severe adverse outcomes such as cardiac arrest, ICU transfer, or in-hospital mortality \cite{nemati2018interpretable,churpek2016multicenter}. Traditional early warning systems rely on manually designed scoring rules derived from physiological measurements. Widely adopted examples include the Modified Early Warning Score (MEWS) \cite{subbe2001validation} and the National Early Warning Score (NEWS) \cite{rcp2012national}, which aggregate a small set of vital signs into interpretable risk scores.

With the increasing availability of electronic health records (EHRs), machine learning methods have significantly improved clinical deterioration prediction. Early approaches applied logistic regression and tree-based models using a broader set of clinical variables such as laboratory results, demographics, and medical history \cite{churpek2016multicenter}. More recent work has explored deep learning models capable of capturing temporal patient trajectories, including recurrent neural networks and transformer-based architectures \cite{rajkomar2018scalable,harutyunyan2019multitask,moor2020predicting,huang2025mimic}. These models can generate individualized risk predictions for adverse outcomes such as sepsis onset, ICU admission, and mortality.

Despite substantial progress in predictive modeling, most existing studies primarily focus on improving patient-level prediction accuracy \cite{shickel2017deep}. Comparatively less attention has been given to how predictive outputs can support operational decision-making at the hospital level \cite{sendak2020presenting,finlayson2021clinician,wiens2019do}.

\subsection{Optimization for Healthcare Resource Planning}
Operations research methods have long been used to support decision-making in healthcare systems, including staff scheduling, bed allocation, and patient flow management \cite{burke2004state,ernst2004staff}. In intensive care units (ICUs), staffing decisions are particularly challenging due to uncertain patient demand and the limited availability of specialized personnel \cite{saville2019how}. Prior studies have proposed optimization models to determine appropriate staffing levels under uncertainty \cite{kim2015two}, typically relying on historical admission patterns, census forecasts, or simplified stochastic demand models \cite{green2007coping}.

However, these operational planning approaches are largely developed independently from modern predictive analytics. Staffing models are usually driven by aggregated historical statistics rather than patient-level risk signals \cite{defraeye2016staffing}. As predictive models become increasingly accurate, integrating real-time risk information into operational planning presents an opportunity for more proactive and adaptive healthcare management \cite{bates2014big}.

\subsection{From Prediction to Decision: Integrating Predictive Models with Operations}
Bridging predictive modeling and operational decision-making has recently emerged as an important research direction in applied machine learning. Predict-then-optimize frameworks and decision-focused learning approaches aim to align predictive models with downstream decision objectives by jointly optimizing prediction and decision components \cite{bertsimas2020predictive,elmachtoub2022smart,shah2022decision}. While these approaches can improve decision quality, they typically require retraining predictive models together with the optimization layer and often rely on differentiable or relaxed optimization formulations.

In many real-world healthcare deployments, predictive models are developed and validated as standalone systems and cannot be easily modified once deployed due to regulatory and operational constraints. As a result, tightly integrated prediction–optimization frameworks may be difficult to apply in practice.

Related prediction-to-decision integration problems have also been studied in engineering domains such as predictive maintenance, where probabilistic failure predictions are incorporated into maintenance scheduling decisions \cite{depater2022alarm,vanderhaar2024predict}. These studies demonstrate how predictive uncertainty can inform operational planning in safety-critical systems.

However, the integration of predictive signals into healthcare operational planning remains relatively underexplored. In particular, methods for translating patient-level risk predictions into population-level demand signals suitable for operational decision models are still limited. Addressing this challenge is essential for enabling predictive models to support real-world healthcare resource planning.

This work addresses this gap by proposing an uncertainty-aware framework that aggregates patient-level deterioration predictions and integrates them into ICU staffing decision models. The following section introduces the proposed experimental and decision framework in detail.

\section{Uncertainty-Aware Risk Aggregation Framework}

Predictive models for clinical deterioration provide patient-level risk estimates, yet hospital staffing decisions will be made at the ICU level under substantial operational uncertainty. Translating heterogeneous patient risks into reliable workload signals therefore requires an aggregation mechanism that accounts for predictive uncertainty and operational constraints.

\begin{figure}[t]
\centering
\includegraphics[width=0.9\linewidth]{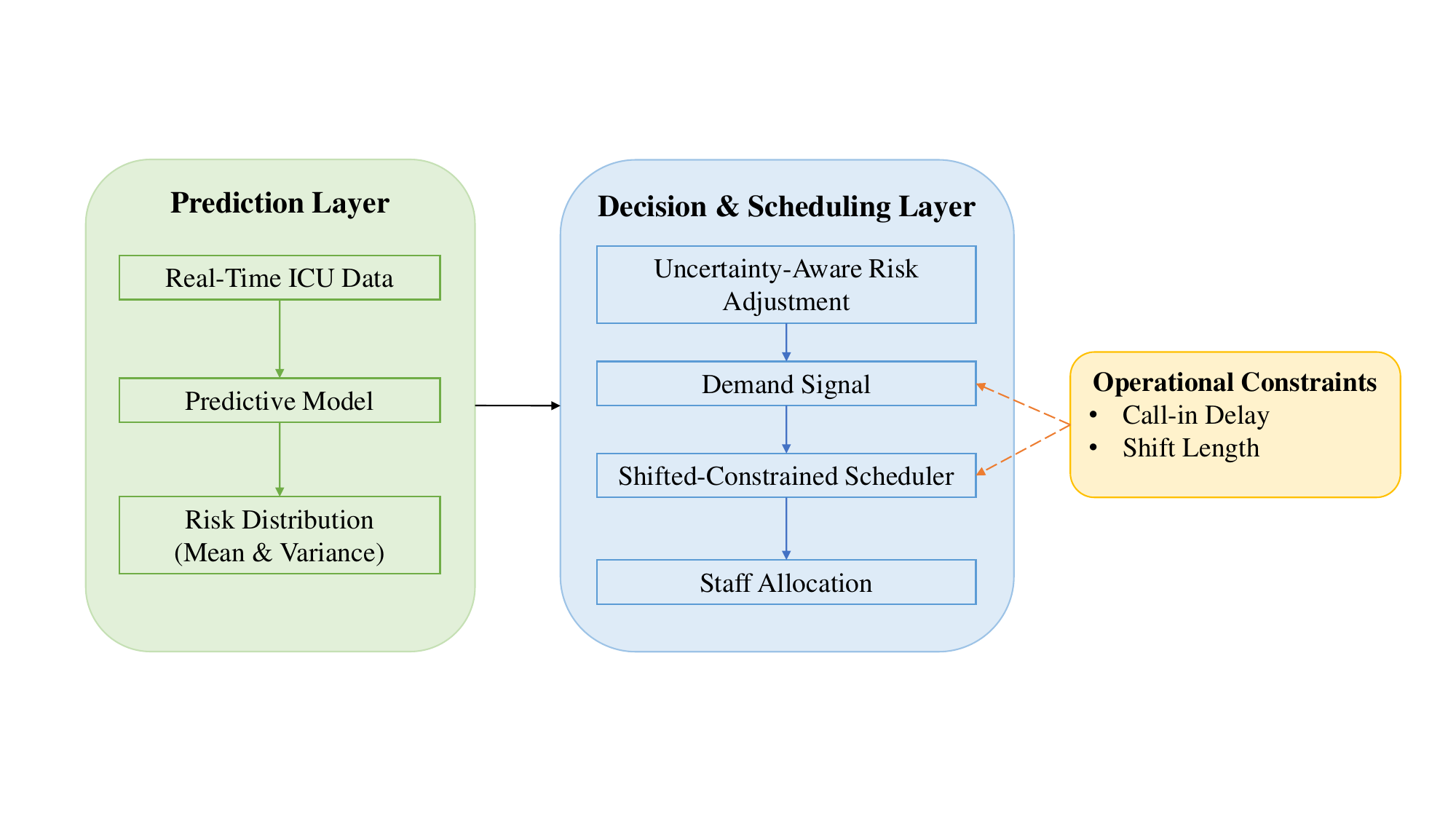}
\caption{Overview of the uncertainty-aware risk aggregation framework}
\label{fig:framework}
\end{figure}

We propose an uncertainty-aware risk aggregation framework that converts patient-level deterioration probabilities into staffing demand estimates and integrates them with delay-constrained scheduling decisions. As illustrated in Fig.~\ref{fig:framework}, the framework consists of three layers: (i) patient-level risk prediction from EHR data, (ii) uncertainty-aware aggregation that constructs population-level demand signals, and (iii) an operational scheduling layer that determines feasible staffing allocations under shift and hiring constraints.

This layered design separates predictive modeling from operational decision-making, allowing validated clinical prediction models to be incorporated into hospital planning systems without retraining.

\subsection{Prediction Layer}
The prediction layer estimates the probability that each patient will require vasopressor treatment within the next 24 hours. We adopt the Vasopressor Requirement (VR) prediction task with treatment variables introduced by \cite{huang2025mimic}. For patient $i$ at time $t$, the model outputs a deterioration probability,

\begin{equation}
p_{i,t} = P(y_{i,t+24}=1 \mid H_{i,t}),
\end{equation}
where $y_{i,t+24}$ denotes a binary indicator of vasopressor usage and $H_{i,t}$ represents the observed clinical history, including physiological measurements, laboratory results, treatment variables, and demographic information. We adopt the benchmark's LSTM-based predictor, producing a sequence of patient-level risk estimates $\{p_{i,t}\}$.

\subsection{Risk Aggregation and Demand Estimation}
 We introduce a risk aggregation mechanism that translates 24-hour patient-level deterioration probabilities into a high-frequency operational demand signal to enable staffing decisions to be made at the ICU level rather than the individual patient level.

The original risk predictions $\{p_{i,k}\}$ are generated at 4-hour intervals, and are resampled to an hourly resolution via linear interpolation. To prevent highly volatile scheduling directives and abrupt staff fluctuations, these initial hourly prediction trajectories are then stabilized using a rolling window average,
\begin{equation}
\bar{p}_{i,t} = \frac{1}{W}\sum_{k=\max(0, t-W+1)}^{t} p_{i,k},
\end{equation}
where $W$ denotes the length of the smoothing window, set to 6 hours to align with our anticipatory scheduling horizon. 

Second, a nonlinear transformation is applied to map this smoothed probability into a risk-calibrated operational demand signal,
\begin{equation}
\tilde{p}_{i,t} = \bar{p}_{i,t}^{\gamma}.
\end{equation}
The exponent $\gamma$ controls how sharply patient-level risk scores are translated into operational demand signals. With $\gamma>1$, low-confidence fluctuations are attenuated, while relatively high-risk trajectories remain influential in the aggregate signal. This mapping is not intended to uniformly inflate individual probabilities. Operational conservatism is introduced through the dynamic uncertainty buffer in Eq.~(5), which accounts for the dispersion of the aggregated demand distribution. In our configuration, $\gamma$ is fixed to $1.2$ on the validation set. With these risk-calibrated signals defined at the patient level, we proceed to aggregate them into a total ICU-level staff requirement.

Aggregated demand is constructed through a mean-variance approximation. Let $S_t$ denote the random aggregate demand at time $t$, defined as the sum of individual patient deterioration events, $S_t = \sum_{i=1}^{N} X_{i,t}$, where each patient's outcome follows a Bernoulli distribution $X_{i,t} \sim \text{Bernoulli}(\tilde{p}_{i,t})$. Although patient outcomes may exhibit weak correlations, assuming conditional independence provides a tractable approximation that is commonly used in aggregate demand modeling. Under this assumption, the expectation $\mu_t = \mathbb{E}[S_t]$ and variance $\sigma_t^2 = \text{Var}(S_t)$ of the aggregate demand are given by:
\begin{equation}
\mu_t = \sum_i \tilde{p}_{i,t}, \qquad
\sigma_t^2 = \sum_i \tilde{p}_{i,t}(1-\tilde{p}_{i,t}).
\end{equation}

To incorporate predictive uncertainty, the operational demand estimate $D_t$ is constructed using a mean-variance adjustment:
\begin{equation}
D_t = \left\lceil \mu_t + \zeta \sigma_t \right\rceil,
\end{equation}
where $\zeta$ controls the desired confidence level to satisfy a specific target service requirement. The ceiling function guarantees integer staffing requirements. This formulation propagates patient-level predictive uncertainty into a population-level workload estimate used for staffing decisions.

\subsection{Operational Staffing Model}
Given the hourly demand sequence $D_t$, staffing decisions should respect operational constraints such as hiring delays and fixed shift lengths. Let $x_t$ denote the number of nurses requested at decision time step $t$. Due to an advance notice requirement of $\tau$ hours, these nurses will begin their shifts at time $t+\tau$ and work for a fixed duration of $L$ hours.

Let $k$ denote the time at which a requested nurse starts a shift. Since each request must be submitted $\tau$ hours in advance, the nurses who start at time $k$ correspond to the earlier staffing decision $x_{k-\tau}$. Because each shift lasts for $L$ hours, the active staff at time $t$ consists of all nurses whose shift start times fall between $t-L+1$ and $t$. Therefore, the number of active nurses at time $t$ is

\begin{equation}
s_t = \sum_{k=t-L+1}^{t} x_{k-\tau}.
\end{equation}

The realized demand $D_t^{real}$ may deviate from predictions due to stochastic patient outcomes. Shortage $u_t$ and overstaffing $o_t$ are defined as

\begin{equation}
u_t = \max(0, D_t^{real}-s_t), \qquad
o_t = \max(0, s_t-D_t^{real}).
\end{equation}

Staffing decisions balance operational cost and service reliability. The minimal total cost over a planning horizon $T$ is

\begin{equation}
\min_{x_t} \sum_{t=1}^{T}  c_s x_t + c_u u_t + c_o o_t,
\end{equation}

where $c_s$, $c_u$, and $c_o$ represent staffing, shortage, and overstaffing costs, respectively. In ICU environments, shortage penalties are typically much higher than overstaffing costs ($c_u \gg c_o$), reflecting the importance of maintaining adequate clinical coverage. The optimization objective uses linear shortage and overstaffing costs; CVaR is used in the evaluation stage to quantify the tail severity of residual understaffing rather than as an explicit constraint in Eq.~(8).

Solving a global mixed-integer program at every time step is impractical in real-time hospital environments. We therefore adopt a rolling forward-scheduling policy. At any decision time step $t$, we calculate the projected baseline staffing level at time $t+\tau$ (prior to the current hiring decision $x_t$ and $\tau<6$) as $\hat{s}_{t+\tau}$. The algorithm then requests additional staff $x_t$ whenever the projected demand $D_{t+\tau}$ exceeds this baseline:

\begin{align}   
        \hat{s}_{t+\tau} &= \sum_{k=t-L+1}^{t-1} x_k,\\
        x_t &= \max(0, D_{t+\tau} - \hat{s}_{t+\tau}).
\end{align}

This rule respects the operational constraints while maintaining computational efficiency for real-time staffing decisions.

\section{Experiments}
\subsection{Dataset and Prediction Backbones}
All experiments are conducted using patient trajectories derived from the MIMIC-IV database (version 3.1) \cite{johnson2023mimic}. Access to the database was granted upon completion of the required CITI human subjects research training, strictly conforming to the PhysioNet credentialed Data Use Agreement. The VR prediction task from the MIMIC-Sepsis benchmark \cite{huang2025mimic} serves as the clinical prediction component.

The LSTM-based backbone processes the extracted physiological measurements, laboratory variables, and treatment indicators to estimate the probability of vasopressor administration within the next 24 hours. Predictive performance is verified on the validation set using the Area Under the Receiver Operating Characteristic curve (AUROC) \cite{fawcett2006introduction} and the Area Under the Precision-Recall Curve (AUPRC) \cite{davis2006relationship}.

To align these predictive outputs with the high-frequency requirements of our scheduling framework, the continuous risk signals are discretized into an hourly operational grid via linear interpolation. These hourly predicted probabilities then serve as the raw input for our uncertainty-aware aggregation layer.

\subsection{Experimental Setup}
We employ a rolling daily operational planning horizon where staffing decisions are updated hourly based on predictive demand distributions and operational constraints. Unless otherwise specified, experiments include $N=30$ patients, a hiring delay $\tau=2$ hours, and a fixed shift length $L=8$ hours. To ensure a fair evaluation and prevent scenario-specific overfitting, all algorithmic tuning parameters are fixed globally across all experiments. The smoothing window is set to $W=6$ hours, the risk-aversion coefficient is $\gamma = 1.2$, and the service-level parameter is $\zeta = 1.28$. 

To evaluate generalization across heterogeneous demand realizations, multiple random seeds are used to sample patient subsets from the database. Each seed produces an independent 24-hour probabilistic demand trajectory. All competing policies are evaluated on identical realizations to ensure paired comparisons and eliminate stochastic bias.

\subsection{Evaluation Metrics}
To assess the operational and risk implications of different staffing policies, performance is evaluated along four dimensions: (i) economic efficiency, (ii) service adequacy, (iii) operational stability, and (iv) tail-risk exposure.

Total economic efficiency over the planning horizon is defined as the aggregate staffing cost $\sum_t c_s x_t$. Service adequacy is measured by the cumulative shortage $\sum_t u_t$ and total overstaffing $\sum_t o_t$. Operational stability is captured by staffing volatility, defined as the absolute shift-to-shift variation in active personnel $\sum_t |s_t - s_{t-1}|$.

Finally, extreme shortage risk is reported using Conditional Value-at-Risk (CVaR):
\begin{equation}
\text{CVaR}_\alpha = \mathbb{E}[u_t \mid u_t \ge \text{VaR}_\alpha].
\end{equation}
where $\text{VaR}_\alpha$ is the Value-at-Risk at confidence level $\alpha$ (satisfying $\mathbb{P}(u_t \le \text{VaR}_\alpha) = \alpha$). In our evaluation, we set tail risks for $\alpha \in \{0.90, 0.95\}$.

In ICU staffing contexts, the consequences of a staffing imbalance are highly asymmetric. Severe understaffing can compromise patient safety and care quality, whereas temporary overstaffing mainly results in financial inefficiency. Consequently, our risk measure strictly focuses on the tail risk of staffing shortages. CVaR emphasizes extreme shortage scenarios and quantifies the expected severity of understaffing beyond a specified confidence level $\alpha$. It is used as an evaluation metric rather than an optimization constraint in our implementation. In contrast, overstaffing does not represent a safety-critical risk and is therefore modeled separately as a linear cost component in the objective function.

\subsection{Baseline Policies}
We compare the proposed framework with four benchmark staffing strategies implemented under identical operational constraints.

\textbf{Deterministic policy.} Staffing decisions are based solely on expected demand $\mu_t$.

\textbf{Reactive policy.} Hiring follows realized demand $D_t^{real}$ with delay $\tau$.

\textbf{Variance-only policy.} Staffing uses a variance-inflated estimate $\mu_t + \zeta\sigma_t$.

\textbf{Oracle policy.} Staffing is determined using realized demand without prediction error or hiring delay.

These baselines allow us to evaluate whether calibrated risk mapping, dynamic variance buffering, and temporal smoothing improve upon expectation-based planning, purely reactive adjustment, and conventional variance-buffering strategies.

\section{Results}

\subsection{Backbone Sensitivity Analysis}
We first examine whether the proposed aggregation framework depends on the specific predictive model used to estimate patient deterioration risk. Three representative backbones are evaluated: logistic regression, an LSTM-based temporal model, and a Transformer-based sequence model. For each backbone, predicted probabilities are aggregated using the proposed uncertainty-aware aggregation mechanism to estimate shift-level demand, followed by the identical staffing decision layer. 

\begin{table}[tb] 
\centering 
\caption{Backbone sensitivity analysis} \label{tab:backbone} 
\resizebox{0.7\textwidth}{!}{%
\begin{tabular}{l|c|c|c|c} 
\hline 
\textbf{Model} & \textbf{AUROC} & \textbf{AUPRC} & \textbf{Cost (Mean)} & \textbf{Shortage (Mean)} \\ 
\hline 
Logistic Regression & 0.8605 & 0.6809 & 23.4 & 15.0 \\ 
LSTM & 0.8419 & 0.6612 & 23.4 & 16.6 \\ 
Transformer & 0.8646 & 0.6890 & 23.1 & 16.2 \\ 
\hline 
\end{tabular}%
}
\end{table}

Table~\ref{tab:backbone} shows that the resulting staffing metrics remain remarkably robust despite variations in predictive accuracy across models. Specifically, mean staffing costs stay highly consistent, and shortage statistics fluctuate only within a narrow operational margin. This indicates the aggregation mechanism is robustly model-agnostic, seamlessly integrating diverse predictive outputs without requiring downstream modifications. Such operational stability is crucial in practical deployments, where predictive models frequently evolve over time while decision pipelines must remain reliable.

\subsection{Comparative Analysis of Staffing Policies}
Table~\ref{tab:main} summarizes the operational performance of the proposed framework against external benchmark policies under realistic MIMIC-IV demand trajectories. 

\begin{table}[tb]
\centering
\caption{Performance comparison across staffing policies}
\label{tab:main}
\resizebox{0.9\textwidth}{!}{%
\begin{tabular}{l|l|l|l|l|c}
\hline
\textbf{Method} & \textbf{Cost} & \textbf{Shortage} & \textbf{Overstaff} & \textbf{Volatility} & \textbf{CVaR (90\%)} \\
\hline
Deterministic & 19.9 $\pm$ 3.2 & 22.9 $\pm$ 13.9 & 26.4 $\pm$ 23.1 & 8.7 $\pm$ 1.7 & 6.49 \\
Reactive & 33.9 $\pm$ 3.1 & 14.4 $\pm$ 4.5 & 120.6 $\pm$ 32.9 & 13.7 $\pm$ 1.4 & 0.60 \\
Variance-only & 27.2 $\pm$ 3.4 & 14.8 $\pm$ 5.0 & 72.4 $\pm$ 32.4 & 11.1 $\pm$ 1.5 & 0.62 \\
Proposed (Full) & 23.4 $\pm$ 3.3 & 16.6 $\pm$ 7.4 & 46.3 $\pm$ 29.8 & 9.8 $\pm$ 1.4 & 4.44 \\
Oracle & 19.3 $\pm$ 5.8 & 0.0 $\pm$ 0.0 & 10.6 $\pm$ 5.8 & 2.4 $\pm$ 1.6 & 0.00 \\
\hline
\end{tabular}%
}
\end{table}

Three key observations emerge. First, the deterministic policy achieves the lowest average cost but exhibits the highest tail risk (CVaR = 6.49), indicating severe vulnerability to demand surges. Second, while the reactive policy substantially reduces this tail risk, it triggers massive overstaffing (120.6 average nurse-hours), resulting in the highest operational cost and volatility due to ex-post corrections. Third, the proposed framework achieves a favorable balance. It reduces cumulative shortage compared to the deterministic baseline while maintaining substantially lower overstaffing than the reactive and variance-only strategies. A paired t-test on seed-level cumulative shortage shows a significant reduction relative to the deterministic baseline ($p<0.001$). Overall, uncertainty-aware anticipation improves system robustness while avoiding excessive capacity inflation.

\subsection{Uncertainty Ablation Study}
To systematically evaluate the impact of each uncertainty-modeling component, we compare four internal variants: (i) Mean-only ($\mu$) serves as the unadjusted foundation by aggregating expected probabilities directly, (ii) Static Margin ($\mu + \sigma_{stat}$) introduces a fixed population-level buffer, (iii) Dynamic Margin ($\mu + \sigma_{dyn}$) estimates variance directly from the real-time predicted risk distribution, and (iv) Full Framework ($\mu + \sigma_{dyn}$) combines calibrated risk mapping, dynamic variance estimation, and temporal smoothing.

\begin{table}[tb]
    \centering
    \caption{Mechanism ablation and tail-risk analysis}
    \label{tab:ablation_main}
    \resizebox{0.95\textwidth}{!}{%
    \begin{tabular}{l|c|c|c|c}
        \hline
        \textbf{Ablation Variant} & \textbf{Cost (Mean)} & \textbf{Shortage (Mean)} & \textbf{P90 Shortage} & \textbf{P95 Shortage} \\
        \hline
        1. Mean-only ($\mu$)                  & 19.9 & 22.9 & 45.1 & 46.5 \\
        2. Static Margin ($\mu + \sigma_{stat}$)  & 26.9 & 14.7 & 21.2 & 23.5 \\
        3. Dynamic Margin ($\mu + \sigma_{dyn}$)  & 22.9 & 16.9 & 29.1 & 30.0 \\
        4. Full Framework ($\mu + \sigma_{dyn}$) & 23.4 & 16.6 & 28.3 & 31.5 \\
        \hline
    \end{tabular}%
    }
\end{table}

\begin{figure}[tb]
\centering
\begin{minipage}[t]{0.48\textwidth}
\centering
\includegraphics[width=\textwidth]{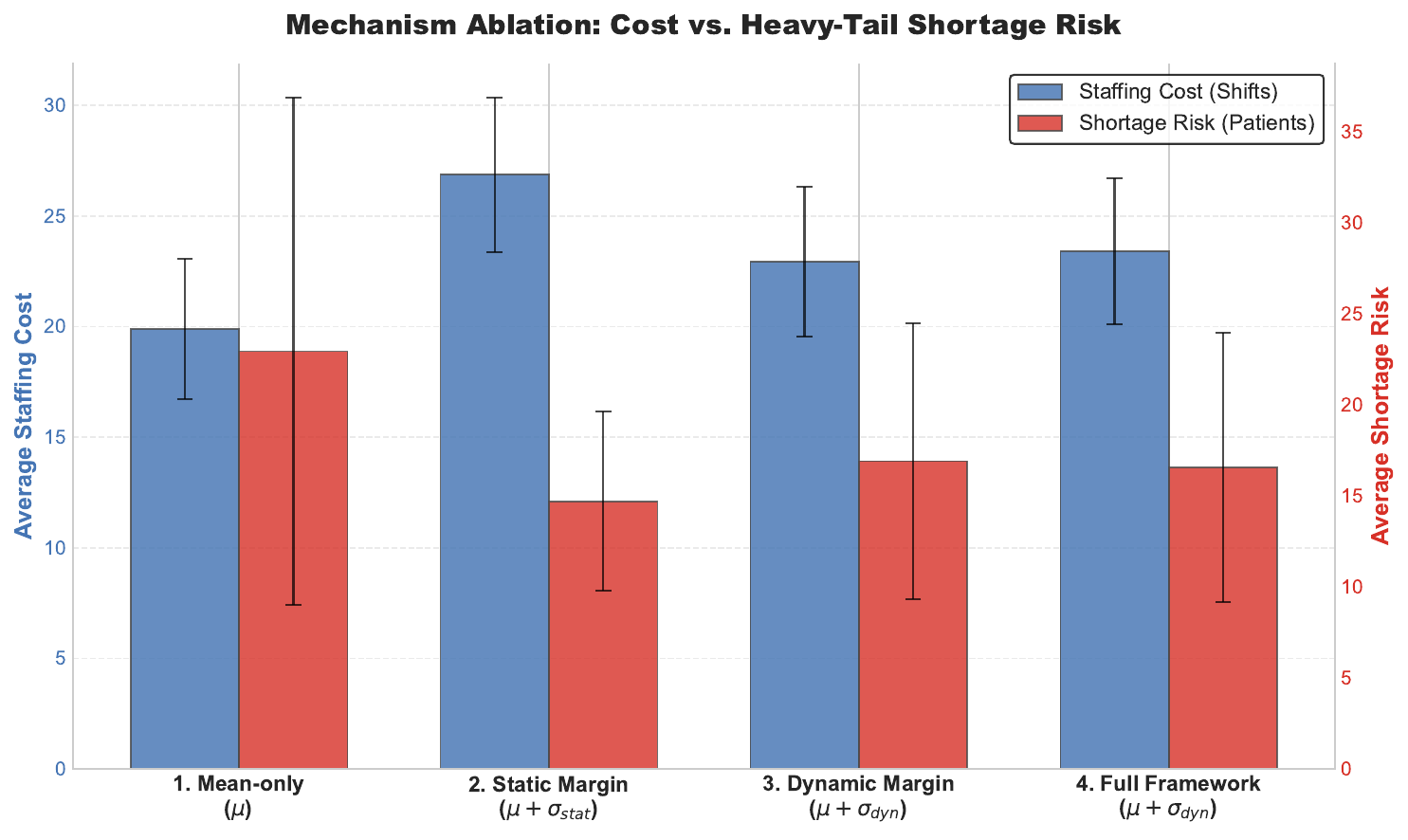}
\caption{Impact of uncertainty modelling on staffing cost and shortage. Bars report mean values across 30 random seeds, with error bars denoting one standard deviation.}
\label{fig:ablationbar}
\end{minipage}
\hfill
\begin{minipage}[t]{0.48\textwidth}
\centering
\includegraphics[width=\textwidth]{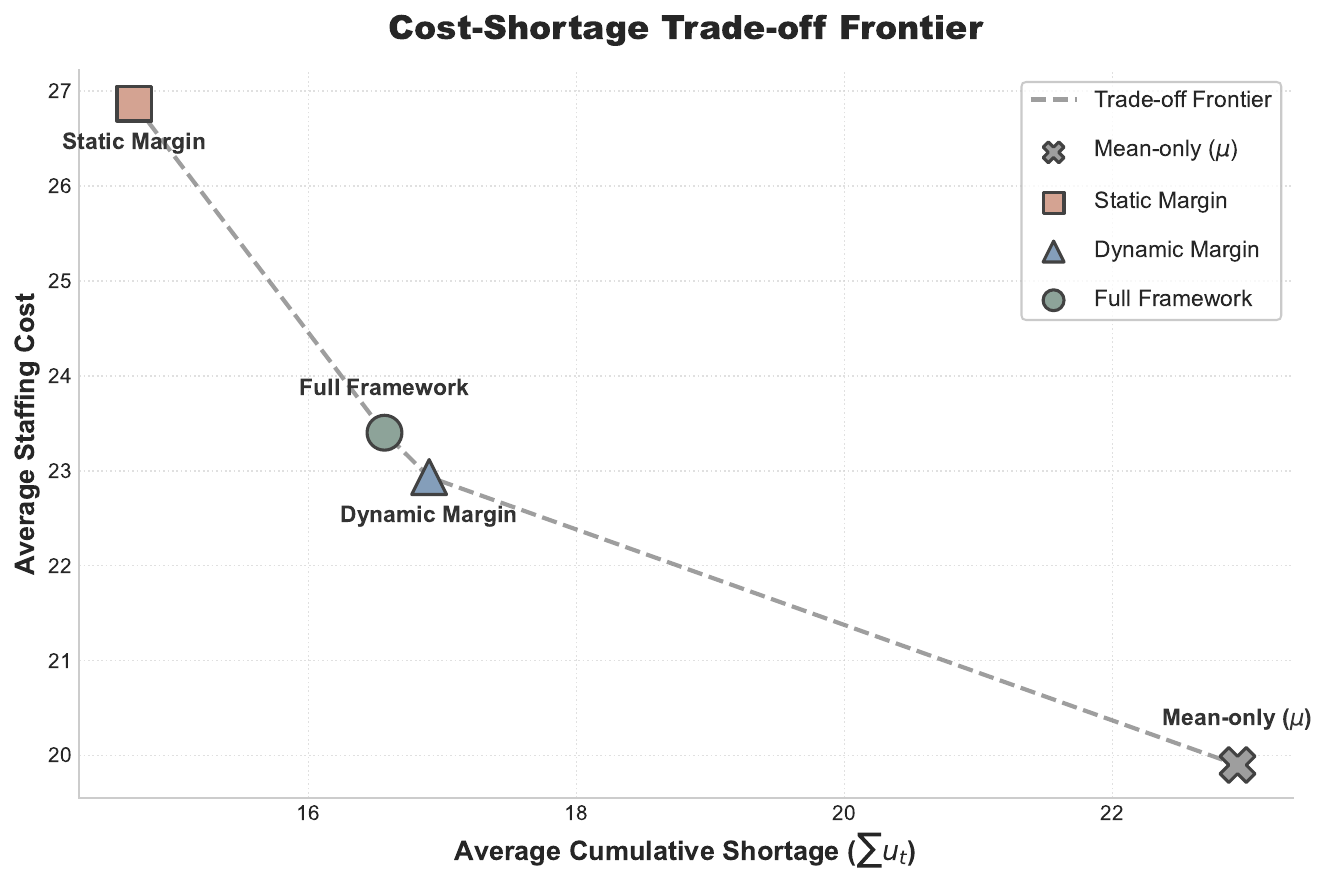}
\caption{Cost-shortage trade-off across uncertainty aggregation strategies. Each point reports the mean across 30 random seeds.}
\label{fig:pareto}
\end{minipage}
\end{figure}

Table~\ref{tab:ablation_main} reports the resulting staffing cost and shortage statistics. The P90/P95 shortage metrics quantify tail-risk exposure, representing the magnitude
of unmet clinical demand during the worst 10\% and 5\% of extreme operational scenarios, respectively. The deterministic approach accumulates low-probability deterioration events, leading to severe shortage risks. Introducing uncertainty estimates stabilizes demand and reduces shortage frequency. The full framework provides a more favorable empirical trade-off by reducing cumulative shortage relative to the mean-only baseline while avoiding the large overstaffing induced by more conservative buffers. Fig.\@ \ref{fig:ablationbar} visualizes the average shortage across variants, while Fig.\@ \ref{fig:pareto} illustrates the resulting cost-shortage trade-off. These results highlight the importance of modeling uncertainty when translating probabilistic predictions into operational decisions.

\subsection{Calibration Stress Test}
Predictive models deployed in practice often experience calibration drift over time, systematically overestimating or underestimating clinical risks due to shifts in patient demographics or clinical protocols. We set this systematic bias by applying a multiplicative scaling factor $\beta \in [0.7, 1.3]$ to the original predictions. The drifted probability is thus given by $\tilde{p}_{i,t} = \min(1, \beta \cdot p_{i,t})$.

\begin{table}[tb]
    \centering
    \caption{Systematic bias robustness stress test}
    \label{tab:stress_test_summary} 
    \resizebox{0.75\textwidth}{!}{%
    \begin{tabular}{l|c|c|c}
        \hline
        \textbf{Scenario} & \textbf{Cost (Mean)} & \textbf{Shortage (Mean)} & \textbf{P90 Shortage} \\
        \hline
        Original & 23.4 & 16.6 & 28.3 \\
        Overestimation (1.3x) & 28.9 & 14.5 & 21.1 \\
        Underestimation (0.7x) & 17.3 & 30.3 & 59.3 \\
        Noisy (+N) & 23.7 & 16.4 & 28.2 \\
        \hline
    \end{tabular}%
    }
\end{table}

\begin{figure}[tb]
\centering
\includegraphics[width=0.45\linewidth]{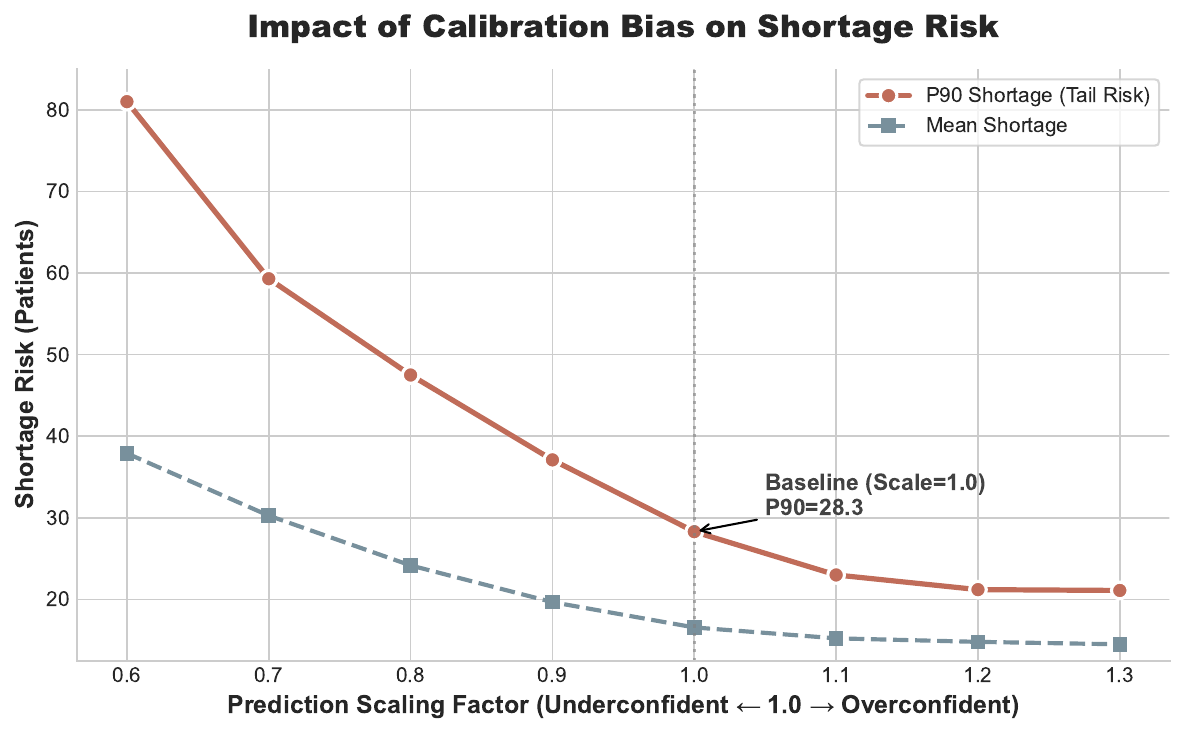}
\caption{Impact of systematic probability drift on shortage risk}
\label{fig:calibration}
\end{figure}

Table~\ref{tab:stress_test_summary} and Fig.\@ \ref{fig:calibration} demonstrate that moderate overestimation (1.3x) slightly increases staffing levels, safely absorbing shortage risk. In contrast, underestimation systematically downplays deterioration risks, exposing the ICU to severe shortages (P90 shortage spikes to 59.3). The framework remains operationally stable under stochastic noise and moderate positive bias, confirming its practical suitability for real-world clinical decision pipelines where predictive drift is common.

\section{Decision-Level Case Analysis}





\subsection{24-Hour ICU Planning Case Study}
To illustrate how the proposed framework supports operational decision-making, a 24-hour ICU staffing scenario is analyzed.

For a selected day in the test dataset, macroscopic patient deterioration risks are first predicted using the trained risk prediction model. These patient-level probabilities are then aggregated into hourly unit-level demand distributions using the uncertainty-aware aggregation mechanism. This step incorporates both backward rolling stabilization and calibrated risk mapping to construct operational signals. Based on these estimated demand trajectories, the rolling-horizon scheduling policy generates the recommended staff level for each decision period under advance notice constraints.

\begin{figure}[htb]
\centering
\includegraphics[width=0.65\linewidth]{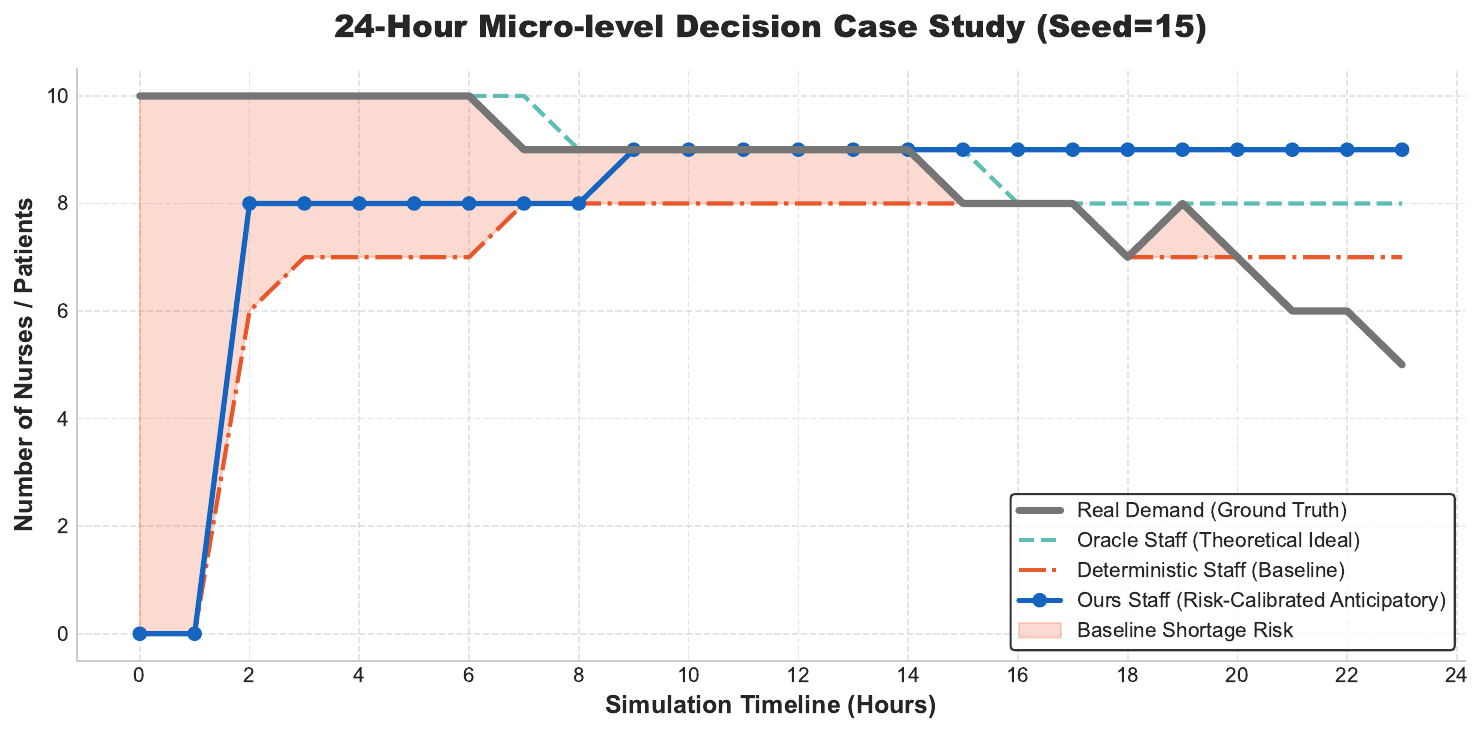}
\caption{24-hour micro-level decision dynamics under MIMIC-IV-derived ICU demand}
\label{fig:studycase}
\end{figure} 

Fig.\@ \ref{fig:studycase} visualizes the resulting staffing schedule, comparing our proposed framework against a deterministic baseline and a theoretical oracle. The figure demonstrates how uncertainty-aware aggregation allows the decision framework to proactively allocate additional staff ahead of actual demand spikes. While the deterministic baseline exposes the unit to severe shortage risks during periods of elevated deterioration, our method successfully buffers these fluctuations. This example illustrates how the proposed pipeline translates macroscopic patient risk predictions into safe, actionable staffing decisions in a realistic ICU planning scenario.
\subsection{Comparative Policy Performance under Realistic Demand}
\begin{figure}[htbp]
\centering
\includegraphics[width=0.9\textwidth]{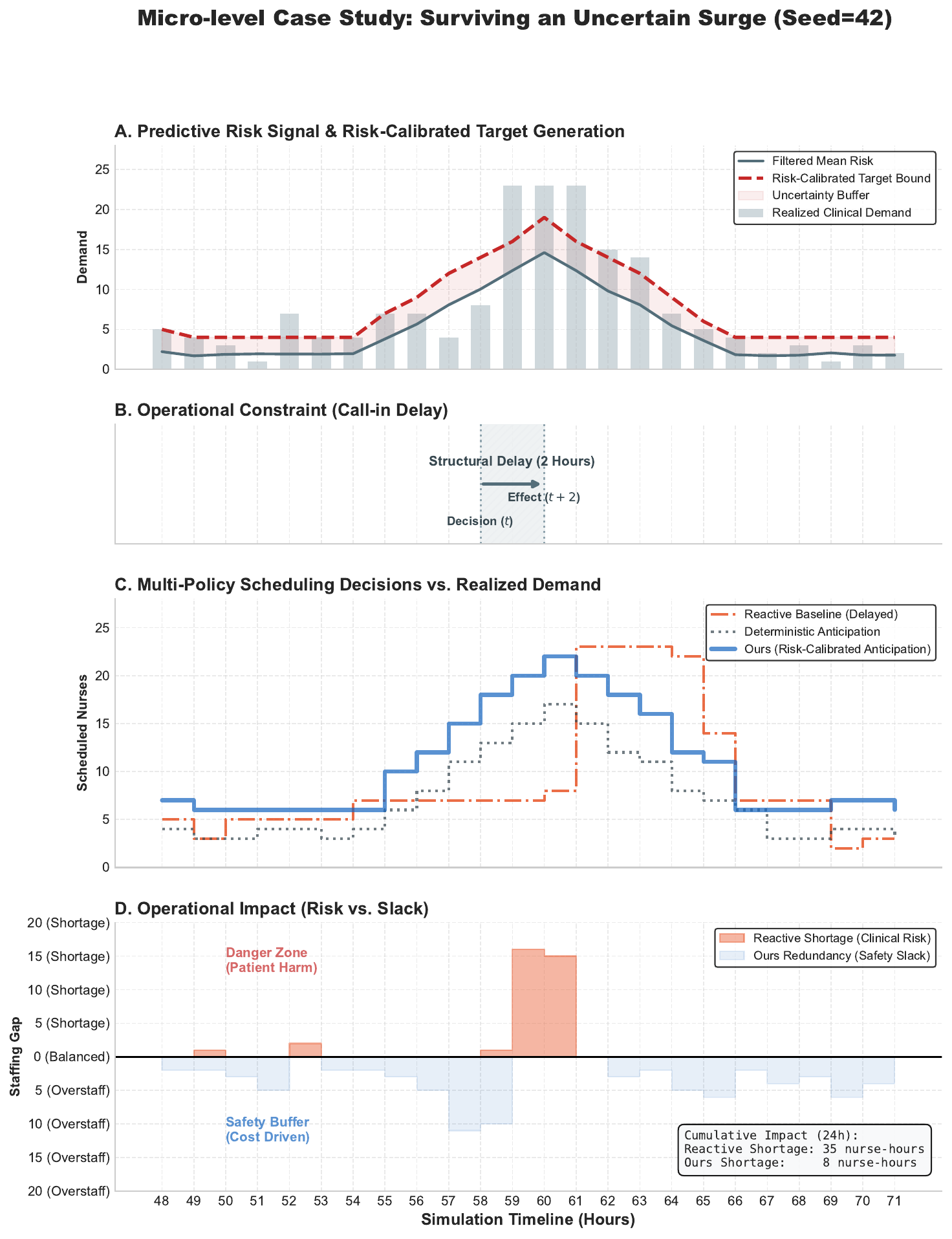}
\caption{Stress-test evaluation under surge conditions (96-hour scenario, seed = 42)}
\label{fig:4panel}
\end{figure}

To evaluate robustness under adverse conditions, we construct a 96-hour synthetic demand trajectory featuring a correlated deterioration surge. Fig.\@ \ref{fig:4panel} presents a representative 24-hour segment of this stress test.

Panel A shows the aggregated predictive risk signal and its adjusted staffing target. The proposed framework increases staffing prior to the realized peak, reflecting anticipatory behavior. Panel B illustrates the two-hour hiring delay ($\tau = 2$), emphasizing that decisions have to be made before demand is fully observed. Under such constraints, purely reactive strategies inherently lag. Panel C compares staffing trajectories. The reactive policy responds only after shortages materialize, leading to delayed corrections. The deterministic baseline anticipates demand trends but underestimates the surge magnitude. In contrast, the proposed method increases staffing earlier and maintains smoother adjustments through the peak period. Panel D quantifies the staffing gaps. Over the displayed horizon, cumulative shortage is reduced from 35 nurse-hours under the reactive policy to 24 nurse-hours under the proposed approach.

These results demonstrate that the advantage of uncertainty-aware anticipation persists under severe surge conditions and is not limited to average-case scenarios.

\section{Conclusions and Limitations}

Empirical evaluation under realistic MIMIC-IV trajectories demonstrates that the proposed approach achieves a favorable cost-shortage trade-off. Compared with deterministic and reactive baselines, the framework reduces shortage exposure and maintains operational stability, even when subjected to systematic predictive bias. Mechanism analyses further illustrate how dynamic uncertainty adjustments safeguard staffing trajectories against extreme demand surges at the micro-level.

The primary contribution of this work is methodological rather than purely predictive. It provides a structured, robust approach for embedding probabilistic clinical information into operational decision pipelines under structural constraints. Although demonstrated within an ICU setting, the core mechanism of adjusting demand signals via calibrated uncertainty to account for mobilization delays extends naturally to other resource allocation problems facing stochastic demand and response lags.

Several limitations guide future work. First, the staffing policy employs a forward sequential decision rule rather than a global mixed-integer optimization model. This prioritizes computational tractability and real-world deployability over theoretical global optimality. Second, our experimental setting assumes a homogeneous staffing pool and aggregate patient demand, abstracting away individual patient acuity levels and nurse-to-patient assignment constraints (e.g., continuity of care and specialized skill matching). Future research should explore the framework's integration with global optimization schedulers, accommodate multi-day scheduling with heterogeneous staff skills and patient-specific acuity rules, and investigate adaptive mechanisms to correct predictive bias in real time. Despite these simplifications, this study bridges the gap between probabilistic clinical prediction and operational ICU staffing.
\section*{Data and Code Availability}
The MIMIC-IV dataset used in this study is available on PhysioNet \cite{johnson2023mimic}: \url{https://physionet.org/content/mimiciv/3.1/}. To facilitate reproducibility, the code is available at \url{https://github.com/jiaya12/uncertainty-aware-icu-staffing}.

\end{document}